\documentclass[11pt]{article}
\usepackage[letterpaper,margin=1in]{geometry}
\usepackage[T1]{fontenc}
\usepackage[utf8]{inputenc}
\usepackage{lmodern}
\usepackage{microtype}
\usepackage{amsmath,amssymb,amsthm}
\usepackage{booktabs,array,longtable,tabularx}
\usepackage{enumitem}
\usepackage{caption}
\usepackage[authoryear,round]{natbib}
\usepackage{setspace}
\usepackage{ragged2e}
\usepackage[hidelinks]{hyperref}
\usepackage{url}
\usepackage{authblk}

\newtheorem{proposition}{Proposition}

\title{Platform Adaptation Under Governance Interventions: Actor Best-Response Modeling and an External Public-Case Benchmark}
\author[1]{Wesley Shu\thanks{Corresponding author: \href{mailto:shu@energeticparadigm.org}{shu@energeticparadigm.org}}}
\author[2]{Peng WEI\thanks{\href{mailto:weipeng2019@swu.edu.cn}{weipeng2019@swu.edu.cn}}}
\affil[1]{The Institute of Energetic Paradigm}
\affil[2]{College of Plant Protection, Southwest University}
\date{}

\hypersetup{
  pdftitle={Platform Adaptation Under Governance Interventions: Actor Best-Response Modeling and an External Public-Case Benchmark},
  pdfauthor={Wesley Shu; Peng WEI},
  pdfsubject={Platform governance, actor adaptation, and public-case benchmark},
  pdfkeywords={platform governance, digital platforms, actor adaptation, moderation burden, strategic gaming, information systems}
}

\begin{document}
\maketitle

\begin{abstract}
Digital platforms govern by changing rules: rankings, monetization thresholds, moderation standards, verification systems, disclosure requirements, appeal processes, and access policies. These interventions are rarely absorbed passively. Creators, sellers, advertisers, moderators, users, developers, and strategic operators adapt to the new reward surface. This paper develops a platform-adaptation model for evaluating governance interventions as transitions in adaptive multi-actor information systems. The model represents actor best response, strategic gaming opportunity, moderation burden, user-incentive movement, enforcement response, externality formation, and downstream platform stability. We evaluate the model on 72 external public platform-governance cases covering media monetization, ranking systems, verification, delivery platforms, marketplaces, app stores, community platforms, and creator ecosystems. Across 9 methods and 648 method-case evaluations, the full platform-adaptation simulator achieves mean adaptation quality of 0.836338, compared with 0.669731 for a risk-register baseline, 0.589457 for causal-loop analysis, 0.492750 for generic governance critique, 0.369492 for engagement-only optimization, and 0.331965 for baseline policy review. Paired comparisons show a win rate of 1.00 against all tested baselines and channel ablations. The contribution is an information-systems theory and measurement framework showing why platform governance evaluation fails when it treats policy rules as static controls rather than interventions into adaptive actor-response fields.
\end{abstract}


\setstretch{1.08}

\section{Introduction}
Digital platforms are governed through rules. A platform changes its ranking algorithm, adjusts monetization eligibility, raises or lowers moderation thresholds, redesigns verification, modifies seller requirements, restricts API access, alters app-store review practices, or introduces new appeal procedures. In each case the rule is not merely a compliance text. It changes the reward surface of the platform. Actors who depend on the platform observe the new constraints and incentives, infer what is now rewarded or penalized, and alter their behavior. A rule that looks sensible at the moment of announcement can therefore produce a different state after creators, sellers, users, moderators, advertisers, developers, and strategic operators respond.

This adaptive response is not peripheral. It is the core control problem of platform governance. A moderation rule changes the workload and ambiguity faced by moderators. A ranking rule changes the strategies that creators and sellers use to become visible. A monetization threshold changes the tradeoff between compliance, opacity, and gaming. A verification system changes trust signals and impersonation incentives. A delivery-platform policy changes driver behavior, customer expectations, and externalized cost. These responses can improve stability, but they can also displace harm, create new loopholes, or increase the burden of enforcement.

Information-systems research is especially well positioned to study this problem because platforms are socio-technical systems, not merely markets or software products. Platform governance joins technical architecture, boundary resources, organizational control, algorithmic management, complementor incentives, moderation work, and network externalities \citep{tiwana2013,gawer2014,eaton2015,wareham2014,henfridsson2013,kellogg2020}. Yet many governance-evaluation methods still reason as if the rule has a direct effect on the outcome. They ask whether a policy targets the right abuse, reduces a visible cost, improves an engagement metric, or names the major risk. Those questions are necessary, but insufficient. The missing question is: \emph{what post-rule actor-response field does the intervention create?}

This paper develops and evaluates a platform-adaptation model for that question. The model treats a governance intervention as a transition operator on a platform state. The immediate policy action is only one component of the transition. The next state depends on actor best responses, gaming opportunities, moderation burden, enforcement capacity, user incentives, externalities, and stability consequences. The paper then evaluates whether methods recover these transition mechanisms using an external public-case benchmark of 72 platform-governance episodes and 648 method-case evaluations.

The paper makes four contributions. First, it provides a theoretical account of platform governance as actor best-response transition control. Second, it formalizes platform adaptation constructs that connect governance interventions to downstream stability: gaming opportunity, moderation burden, user incentive shift, enforcement response, externality risk, and platform-stability consequences. Third, it introduces a reproducible external public-case benchmark for evaluating whether governance methods identify these mechanisms. Fourth, it shows empirically that a full platform-adaptation simulator outperforms baseline policy review, engagement-only optimization, static cost-benefit analysis, generic governance critique, causal-loop analysis, risk-register analysis, and two ablated versions of the model.

The claim is bounded. The benchmark is grounded in public platform-governance cases; it is not a private field experiment, private platform-log analysis, or deployed policy intervention. It therefore does not claim exact prediction of platform outcomes. It evaluates a more specific and practically important capability: whether a governance-evaluation method can identify the actor adaptation channels and downstream platform-stability consequences that a static policy review tends to miss.

\section{Theoretical Background}
\subsection{Digital platforms as governed socio-technical systems}
Digital platforms coordinate interactions among multiple actor groups through technical architecture, rules, interfaces, algorithms, boundary resources, and economic incentives \citep{tiwana2013,ghazawneh2013,eaton2015,ceccagnoli2012,wareham2014}. They are not only two-sided markets; they are governed information infrastructures whose stability depends on participation, complementor investment, user trust, enforcement capacity, and the continuous management of externalities \citep{parker2005,rochet2003,eisenmann2006,dereuver2018}.

The platform literature has emphasized the tension between openness and control. Openness supports innovation, complementor entry, and network effects, but it also increases coordination difficulty, quality variance, and governance burden \citep{boudreau2010,boudreau2012,benlian2015,gawer2014industry}. Control can reduce abuse and preserve quality, but it can also reduce participation or shift costs to dependent actors. This tension is dynamic because platform actors adapt to the governance regime. Boundary resources, APIs, ranking systems, moderation standards, and monetization policies become objects of strategic interpretation and optimization \citep{ghazawneh2013,eaton2015,tiwana2010}.

\subsection{Algorithmic management and rule-mediated control}
Algorithmic management research shows that digital control is often mediated through rankings, scores, ratings, incentives, task allocation, visibility, and automated enforcement \citep{lee2015,kellogg2020,mohlmann2017,rosenblat2016,wood2019}. These controls are powerful because they govern behavior at scale. They are also contested because actors learn the system, interpret its signals, and adapt under uncertainty. Platform governance therefore operates through a feedback relation: the platform modifies rules, actors learn and respond, the platform observes consequences, and governance is revised again.

This feedback relation differs from a conventional policy-effect view. In a static view, the rule is the cause and the observed outcome is the effect. In an adaptive view, the rule changes the incentive environment; actors choose responses; their responses interact; and the observed outcome is a post-response equilibrium or disequilibrium. A platform rule can therefore fail even when it correctly identifies the original problem.

\subsection{Governance failure as adaptation failure}
Platform governance failure often appears as a mismatch between intended rule effect and realized post-response behavior. A rule intended to reduce low-quality content may lead creators to optimize for permissible but low-value signals. A monetization threshold intended to reward reliability may shift strategic effort toward threshold gaming. A verification rule intended to increase trust may create new impersonation incentives. A marketplace ranking change intended to improve relevance may increase seller manipulation. A moderation rule intended to reduce harm may increase appeal burden and reduce enforcement consistency.

These failures share a common structure. The governance method evaluates the policy against the current state, but the platform actually operates in the state produced after actor adaptation. The relevant unit of analysis is therefore not only policy quality, but transition quality. A useful platform-governance method should estimate whether a rule moves the platform toward a stable post-response state.

\section{Theory Development}
\subsection{Core constructs}
We define six constructs for platform-adaptation evaluation.

\textbf{Actor best response} is the expected behavioral adjustment of a platform actor after a governance intervention. Actors include creators, sellers, developers, advertisers, moderators, users, workers, and strategic adversaries. Best response does not imply perfect rationality; it denotes the directional adaptation that becomes locally attractive under the new rule.

\textbf{Gaming opportunity} is the extent to which the intervention creates profitable loopholes, metric substitutions, evasive strategies, or new optimization targets. Gaming opportunity is high when actors can preserve benefits while avoiding the intended constraint.

\textbf{Moderation burden} is the enforcement, review, ambiguity, appeals, and operational load that the intervention shifts onto human or automated moderation systems. It includes both volume and complexity.

\textbf{User-incentive shift} is the expected movement in user attention, trust, reporting, switching, or participation behavior after the rule changes.

\textbf{Externality formation} is the creation or displacement of costs outside the actor who benefits from the rule response. Examples include user harm, seller crowding, creator instability, labor burden, misinformation spillovers, and reduced trust.

\textbf{Platform stability} is the downstream condition of the platform after actor response, enforcement response, and externality formation. It is not equivalent to engagement. Engagement can rise while stability falls.

\subsection{Propositions}
\begin{proposition}[Actor-response gap]
A governance-evaluation method that scores only the policy's stated intent and immediate target will systematically overestimate policy quality when the rule creates profitable post-rule actor responses that are not represented in the evaluation.
\end{proposition}

\begin{proposition}[Gaming displacement]
When a policy constrains a visible behavior but leaves an adjacent reward path open, strategic actors will tend to shift effort toward the adjacent path, producing apparent compliance with reduced platform stability.
\end{proposition}

\begin{proposition}[Moderation-burden relocation]
A governance intervention can reduce visible abuse while increasing moderation ambiguity, appeal load, or enforcement inconsistency. Such interventions can degrade platform stability even when the primary abuse metric improves.
\end{proposition}

\begin{proposition}[Engagement-instability divergence]
Engagement metrics can move in the opposite direction from platform stability when the rule increases attention while also increasing gaming, externality cost, or participant mistrust.
\end{proposition}

\begin{proposition}[Adaptive control advantage]
A method that jointly represents actor best response, gaming opportunity, moderation burden, enforcement response, externality formation, and stability consequences will recover platform-adaptation quality more accurately than methods that represent only visible policy costs, engagement movement, causal loops, or listed risks.
\end{proposition}

\section{Mechanism Logic: From Rule Change to Platform State}
\subsection{Incentive reorientation}
The first mechanism is incentive reorientation. Platform rules do not merely permit or prohibit behavior; they change the relative payoff of observable strategies. A creator who previously optimized for volume may shift toward advertiser-safe language, a seller may shift toward ranking-compatible inventory signals, and a developer may shift toward whichever boundary resource remains available after an access-policy change. The relevant question is not whether the actor agrees with the rule, but whether the rule makes a new strategy attractive enough to be adopted. This is why a governance intervention must be evaluated as a change in the incentive field rather than as an isolated compliance statement.

Incentive reorientation is especially important for digital platforms because the platform often governs through algorithmic visibility and access rather than direct command. The actor does not need full knowledge of the algorithm to adapt. Partial observability is enough. Once actors infer that a signal is rewarded, they experiment, imitate, and diffuse tactics. The post-rule state is therefore shaped by learning and imitation as much as by formal policy.

\subsection{Gaming and adjacent substitution}
The second mechanism is adjacent substitution. A rule may close one route while leaving a neighboring route open. When actors can preserve benefits by moving to an adjacent tactic, the rule may produce apparent compliance without reducing the underlying instability. This is common in ranking systems, recommendation systems, seller markets, creator monetization, review systems, and verification regimes. The platform sees the targeted behavior decline, but the strategic objective survives in a new form.

Adjacent substitution is why static policy review is structurally weak. A static review asks whether the targeted behavior is addressed. An adaptation review asks whether the actor objective has been neutralized, displaced, or made harder to detect. The distinction matters because many platforms have limited visibility into the full actor response space. A rule that reduces one visible abuse channel may increase the value of a less visible channel, thereby shifting enforcement cost into a region where moderation is slower or less consistent.

\subsection{Moderation burden as endogenous cost}
The third mechanism is moderation-burden relocation. Many governance interventions create work. They generate appeals, ambiguous edge cases, borderline classifications, manual review queues, user reports, and consistency problems. This burden is not merely an implementation cost; it changes the stability of the platform. A rule that cannot be enforced consistently may teach actors that the enforcement regime is noisy. A rule that creates high appeal burden may slow resolution and damage trust. A rule that requires difficult interpretation may produce inconsistent treatment across user groups or domains.

Treating moderation burden as endogenous changes the evaluation problem. The platform cannot evaluate only whether the rule is normatively desirable or whether it targets the right category of harm. It must evaluate whether the enforcement system can carry the burden generated by the actor-response field. If moderation capacity is below the burden created by the rule, the governance intervention can create instability even when its policy intent is sound.

\subsection{Externality displacement}
The fourth mechanism is externality displacement. Platforms often reduce one cost by moving it to another actor group. A seller-policy change can reduce buyer risk while increasing seller compliance cost. A monetization rule can increase advertiser confidence while increasing creator income volatility. A ranking change can improve user relevance while increasing strategic manipulation among creators. A verification policy can reduce friction for some legitimate actors while increasing impersonation risk elsewhere. The platform-level outcome depends on whether the displaced cost is absorbed, amplified, or converted into distrust.

Externality displacement is central to information-systems theory because it shows that the platform is a coupled socio-technical system. A local optimization does not remain local. Costs move through technical architecture, market incentives, user expectations, and organizational processes. A useful evaluation method must therefore ask who pays for the intervention, who learns to exploit it, and which group exits, reduces participation, or changes behavior as a result.

\subsection{Stability as post-response viability}
The fifth mechanism is post-response viability. Platform stability is not the same as policy compliance, engagement, or short-term metric improvement. Stability means that the platform can continue to coordinate actor groups without excessive gaming, externality accumulation, moderation overload, or trust erosion. A stable rule may reduce engagement in the short term if it removes low-quality incentives. An unstable rule may increase engagement by rewarding conflict, attention gaming, or strategic participation.

This distinction motivates the benchmark's primary metric. Overall platform-adaptation quality rewards methods that identify the transition mechanisms shaping the post-response state. The score is not designed to reward rhetorical sophistication. It rewards accurate identification of who adapts, how gaming becomes profitable, where moderation burden moves, what externalities form, how enforcement responds, and whether the downstream state is stable.

\section{Platform-Adaptation Model}
Let $e_t$ denote the platform state before a governance intervention. Let $a_t$ denote the intervention, such as a ranking change, monetization rule, moderation threshold, verification policy, seller requirement, API rule, or appeal procedure. Let $\mathcal{I}$ denote the set of platform actor groups. For actor group $i \in \mathcal{I}$, the post-rule response is represented as
\[
B_{i,t+1}=BR_i(a_t,e_t;G_t,M_t,U_t,X_t,E_t),
\]
where $G_t$ is gaming opportunity, $M_t$ is moderation burden, $U_t$ is user-incentive movement, $X_t$ is externality risk, and $E_t$ is enforcement response. The next platform state is
\[
e_{t+1}=F(e_t,a_t,B_{1,t+1},\ldots,B_{n,t+1},G_t,M_t,U_t,X_t,E_t,\epsilon_t).
\]
Platform stability is scored as a downstream state property:
\[
S_{platform} = W_u U_{user} + W_c U_{creator} + W_p U_{platform} - C_{gaming} - C_{externality} - C_{moderation} - D_{instability}.
\]
This representation is intentionally operational. It does not assume that all platform consequences can be reduced to a single utility function. Instead, it forces the evaluation method to state which response channels are expected to change and how those changes affect stability.

\subsection{Evaluation algorithm}
\begin{enumerate}[leftmargin=*]
\item Identify the governance intervention and affected actor groups.
\item Encode the pre-intervention platform state: dependency, visibility, enforcement capacity, moderation ambiguity, and externality exposure.
\item Estimate actor best responses under the new rule.
\item Estimate gaming opportunity, moderation burden, user-incentive movement, enforcement response, and externality formation.
\item Score the expected post-response platform state.
\item Compare the score against baseline governance methods under the locked rubric.
\end{enumerate}

\begin{figure}[ht]
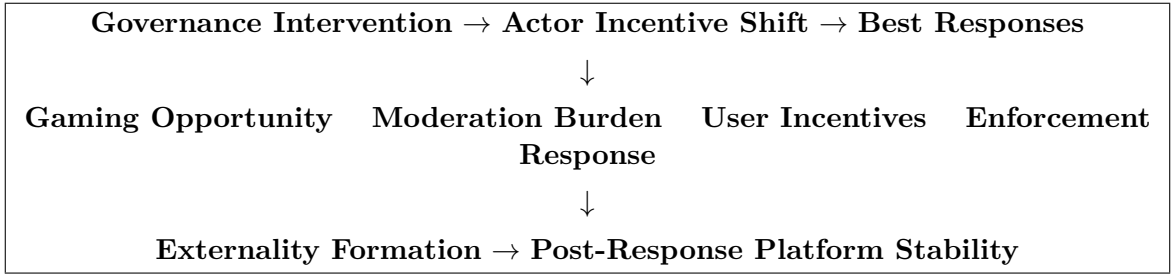

\begin{singlespace}
\centering
\fbox{\begin{minipage}{0.92\linewidth}
\centering
\textbf{Governance Intervention} $\rightarrow$ \textbf{Actor Incentive Shift} $\rightarrow$ \textbf{Best Responses}\\[0.4em]
$\downarrow$ \\[0.4em]
\textbf{Gaming Opportunity} \quad \textbf{Moderation Burden} \quad \textbf{User Incentives} \quad \textbf{Enforcement Response}\\[0.4em]
$\downarrow$ \\[0.4em]
\textbf{Externality Formation} $\rightarrow$ \textbf{Post-Response Platform Stability}
\end{minipage}}
\caption{Platform-adaptation evaluation pipeline. A governance intervention is evaluated by its post-response transition, not only by its stated target.}
\label{fig:pipeline}
\end{singlespace}
\end{figure}

\section{Research Design}
\subsection{External public-case benchmark}
The benchmark contains 72 external public cases. Each case is grounded in a public platform-governance episode and combines a source episode with a policy-change type such as ranking change, monetization rule, moderation threshold, verification policy, marketplace rule, app-store access rule, or appeal process. The benchmark covers media monetization, media ranking, social verification, delivery platforms, digital marketplaces, marketplace ranking, community platforms, creator platforms, and related platform-governance domains.

The benchmark does not use synthetic cases for the main empirical claim. Public sources provide the episode grounding, while the locked rubric specifies what counts as accurate adaptation reasoning. The source manifest is included in the package. The full appendix summarizes the 72 cases by domain, policy type, expected best response, and expected failure mode.

\subsection{Compared methods}
Nine methods are evaluated: baseline platform-policy review, engagement-only optimizer, generic governance critique, static cost-benefit review, risk-register baseline, causal-loop baseline, platform-adaptation simulator without the strategic-gaming channel, platform-adaptation simulator without the moderation-burden channel, and the full platform-adaptation simulator. The set intentionally includes both weak and strong baselines. The risk-register baseline is important because it represents a plausible governance practice: naming hazards and mitigations. The causal-loop baseline is important because it represents feedback reasoning without the full actor-response channel structure.

\subsection{Rubric and scoring}
Each method-case output is scored on creator-adaptation accuracy, moderation-burden accuracy, user-incentive accuracy, engagement-dynamics accuracy, strategic-gaming accuracy, externality-prediction accuracy, enforcement-response accuracy, platform-stability accuracy, overall platform-adaptation quality, and control efficiency. The primary metric is mean overall platform-adaptation quality. Secondary metrics diagnose which channels produce differences among methods.

Construct validity is protected by three design choices. First, the rubric separates adaptation channels rather than collapsing all quality into a vague governance score. Second, ablations remove individual channels, allowing the test to identify whether strategic gaming and moderation burden add explanatory power. Third, paired comparisons evaluate within-case differences, reducing the risk that method rankings are artifacts of case mix.

\subsection{Case construction and source grounding}
The external cases were constructed to avoid two common weaknesses in platform-governance evaluation. The first weakness is purely synthetic scenario design, where the researcher controls both the case and the expected answer. The second weakness is anecdotal case selection, where examples are chosen because they fit the preferred theory. The benchmark instead uses public platform-governance episodes and converts them into repeated policy-change instances under a fixed case schema. Each row includes a public source basis, domain, policy-change type, adaptation shock, actor-dependency indicators, moderation burden, gaming opportunity, externality risk, enforcement probability, expected best response, and expected failure mode.

The purpose of this design is not to claim that public cases provide the same evidentiary depth as internal platform logs. They do not. The purpose is to create an externally grounded test bed where multiple governance-evaluation methods face the same cases, same rubric, same scoring dimensions, and same validation script. This makes the comparison reproducible. It also makes the claim falsifiable: if a method cannot recover expected adaptation channels in public cases, it is unlikely to be reliable in more complex private operational settings.

\subsection{Why the baseline set matters}
The baseline set is chosen to represent distinct governance logics. Baseline policy review represents the minimal static interpretation of a platform rule. Engagement-only optimization represents the common managerial error of treating visible attention as policy success. Static cost-benefit review represents a more formal but still comparatively non-adaptive evaluation style. Generic governance critique represents broad concern without mechanism-specific transition modeling. Causal-loop analysis represents feedback awareness. Risk-register analysis represents structured hazard naming. The two ablations test whether the full model's advantage depends specifically on strategic-gaming and moderation-burden channels.

This structure creates a ladder of increasingly sophisticated governance reasoning. If the full model only outperformed the weakest baseline, the result would be uninteresting. The key result is that the full model outperforms risk-register analysis, causal-loop analysis, and channel ablations. That pattern suggests that the advantage is not simply more words, more caution, or more feedback language. It comes from representing the coupled transition from rule change to actor response to platform state.

\subsection{Statistical treatment}
The benchmark reports aggregate means, bootstrap 95 percent confidence intervals using 2,000 resamples, paired differences, paired t statistics, domain summaries, and failure analysis. The package includes row-level method scores, action scores, bootstrap intervals, source manifest, domain summary, failure analysis, runner, and validator.

\section{Results}
\subsection{Aggregate results}
Table~\ref{tab:aggregate} reports the main aggregate benchmark results. The full platform-adaptation simulator achieves the highest overall platform-adaptation quality and the highest scores on creator adaptation, strategic gaming, moderation burden, and platform-stability accuracy.

\begin{table}[ht]
\begin{singlespace}
\centering
\scriptsize
\caption{External public-case benchmark aggregate results.}
\label{tab:aggregate}
\begin{tabular}{lrrrrrr}
\toprule
Method & Cases & Overall & Creator & Gaming & Moderation & Stability \\
\midrule
Full platform-adaptation simulator & 72 & 0.836338 & 0.857607 & 0.856807 & 0.854240 & 0.793115 \\
No moderation-burden channel & 72 & 0.729647 & 0.827756 & 0.826956 & 0.610945 & 0.549820 \\
No strategic-gaming channel & 72 & 0.716135 & 0.827179 & 0.590129 & 0.823813 & 0.526438 \\
Risk-register baseline & 72 & 0.669731 & 0.641012 & 0.713245 & 0.732419 & 0.651604 \\
Causal-loop baseline & 72 & 0.589457 & 0.597036 & 0.561924 & 0.586646 & 0.592449 \\
Generic governance critique & 72 & 0.492750 & 0.485318 & 0.467397 & 0.500529 & 0.501272 \\
Static cost-benefit review & 72 & 0.440505 & 0.432070 & 0.399090 & 0.475792 & 0.450715 \\
Engagement-only optimizer & 72 & 0.369492 & 0.441623 & 0.240894 & 0.311165 & 0.280944 \\
Baseline policy review & 72 & 0.331965 & 0.334886 & 0.300503 & 0.367864 & 0.327728 \\
\bottomrule
\end{tabular}
\end{singlespace}
\end{table}

The full model outperforms the risk-register baseline by 0.166607 mean quality points and the causal-loop baseline by 0.246881. The gap against engagement-only optimization is 0.466846. This finding supports the proposition that visible engagement is not a reliable substitute for platform-stability evaluation.

\subsection{Paired comparisons}
Table~\ref{tab:paired} reports paired comparisons. The full model wins every paired comparison over the listed baselines and ablations across all 72 cases.

\begin{table}[ht]
\begin{singlespace}
\centering
\scriptsize
\caption{Paired comparisons against the full platform-adaptation simulator.}
\label{tab:paired}
\begin{tabular}{lrrrrr}
\toprule
Comparison & Mean gain & Win rate & t statistic & Min & Max \\
\midrule
Full model vs baseline platform policy review & 0.504374 & 1.00 & 668.16 & 0.489055 & 0.517360 \\
Full model vs engagement only optimizer & 0.466846 & 1.00 & 667.45 & 0.451755 & 0.478433 \\
Full model vs generic governance critique & 0.343588 & 1.00 & 467.26 & 0.331810 & 0.355264 \\
Full model vs static cost benefit & 0.395834 & 1.00 & 601.01 & 0.382143 & 0.405996 \\
Full model vs risk register baseline & 0.166607 & 1.00 & 290.94 & 0.155235 & 0.177158 \\
Full model vs causal loop baseline & 0.246881 & 1.00 & 385.59 & 0.234569 & 0.258390 \\
Full model vs ep no gaming channel & 0.120203 & 1.00 & 102.71 & 0.098762 & 0.139968 \\
Full model vs ep no moderation channel & 0.106691 & 1.00 & 88.88 & 0.085414 & 0.122510 \\
\bottomrule
\end{tabular}
\end{singlespace}
\end{table}

The ablation results are theoretically informative. Removing the strategic-gaming channel reduces quality by 0.120203. Removing the moderation-burden channel reduces quality by 0.106691. These are not cosmetic features. They are core mechanisms through which governance interventions succeed or fail.

\subsection{Bootstrap confidence intervals}
\begin{table}[ht]
\begin{singlespace}
\centering
\scriptsize
\caption{Bootstrap 95 percent confidence intervals for overall platform-adaptation quality.}
\label{tab:bootstrap}
\begin{tabular}{lrr}
\toprule
Method & Mean & 95 percent CI \\
\midrule
Full platform-adaptation simulator & 0.836338 & [0.834914, 0.837822] \\
No moderation-burden channel & 0.729647 & [0.726655, 0.732904] \\
No strategic-gaming channel & 0.716135 & [0.713204, 0.718859] \\
Risk-register baseline & 0.669731 & [0.668365, 0.671116] \\
Causal-loop baseline & 0.589457 & [0.588078, 0.590901] \\
Generic governance critique & 0.492750 & [0.491057, 0.494391] \\
Static cost-benefit review & 0.440505 & [0.439047, 0.441987] \\
Engagement-only optimizer & 0.369492 & [0.367944, 0.371069] \\
Baseline policy review & 0.331965 & [0.330286, 0.333587] \\
\bottomrule
\end{tabular}
\end{singlespace}
\end{table}

The full model's confidence interval is separated from those of the risk-register baseline, causal-loop baseline, generic governance critique, static cost-benefit review, engagement-only optimizer, and baseline policy review. The intervals are also separated from the two channel ablations.

\subsection{Domain-level results}
\begin{table}[ht]
\begin{singlespace}
\centering
\scriptsize
\caption{Domain-level performance summary.}
\label{tab:domain}
\begin{tabular}{lrrrr}
\toprule
Domain & Full model & Risk register & Causal loop & Engagement only \\
\midrule
community platform & 0.836131 & 0.672008 & 0.586358 & 0.375220 \\
creator monetization & 0.834768 & 0.669516 & 0.589766 & 0.369750 \\
creator platform & 0.835802 & 0.670249 & 0.589864 & 0.371126 \\
delivery platform & 0.837464 & 0.670163 & 0.583123 & 0.371292 \\
digital marketplace & 0.841829 & 0.670968 & 0.589935 & 0.370444 \\
marketplace ranking & 0.836930 & 0.669234 & 0.589348 & 0.368711 \\
media monetization & 0.829034 & 0.666352 & 0.585832 & 0.370427 \\
media ranking & 0.837550 & 0.668010 & 0.591326 & 0.371851 \\
mobile ecosystem & 0.832219 & 0.664655 & 0.589514 & 0.366041 \\
mobile marketplace & 0.836762 & 0.671350 & 0.590495 & 0.371385 \\
mobility platform & 0.837237 & 0.667588 & 0.590559 & 0.370079 \\
publishing platform & 0.841213 & 0.672467 & 0.591953 & 0.370432 \\
reviews marketplace & 0.838604 & 0.669899 & 0.590066 & 0.367415 \\
search ranking & 0.840229 & 0.669545 & 0.589208 & 0.366771 \\
short-video ranking & 0.830849 & 0.666594 & 0.588132 & 0.371018 \\
social incentives & 0.832426 & 0.673250 & 0.585644 & 0.368829 \\
social moderation & 0.835928 & 0.668447 & 0.591484 & 0.365599 \\
social ranking & 0.838466 & 0.675541 & 0.595645 & 0.367728 \\
social verification & 0.835143 & 0.670228 & 0.591920 & 0.369624 \\
travel marketplace & 0.834311 & 0.667766 & 0.587547 & 0.367845 \\
\bottomrule
\end{tabular}
\end{singlespace}
\end{table}

The domain results show that the full model's advantage is not driven by a single platform type. The same mechanism appears across creator monetization, media ranking, marketplaces, delivery platforms, community governance, and other platform settings.

\section{Construct Validity and Robustness}
A benchmark for platform governance must answer a construct-validity question: why does the score measure platform adaptation rather than general rhetorical quality? The rubric addresses this problem by assigning separate scores to observable adaptation channels. A method can receive credit for naming a risk without receiving credit for identifying actor best response. It can receive credit for describing engagement effects without receiving credit for strategic-gaming prediction. It can receive credit for describing a moderation rule without receiving credit for anticipating moderation burden. This separation matters because governance failure often arises when one visible policy target improves while another hidden channel deteriorates.

The comparison design also protects against false confidence. The risk-register baseline is not a straw man; it performs better than generic governance critique, static cost-benefit analysis, engagement-only optimization, and baseline policy review. The causal-loop baseline is also meaningful because it captures feedback reasoning. The full model's advantage over these baselines therefore indicates that actor best response, gaming, moderation burden, enforcement response, and externality channels add measurement value beyond generic caution or feedback language.

The limitations are equally important. Public cases provide credible external grounding, but they do not provide private platform logs or controlled deployment outcomes. The locked rubric improves reproducibility, but future work should add independent expert annotation, inter-rater reliability, and prospective validation on new governance interventions. The current evidence supports method-level construct recovery, not exact forecasting of future platform outcomes.

\section{Theoretical Contributions to Information Systems Research}
This paper contributes to information-systems research in three ways.

First, it reframes platform governance as adaptive transition control. Prior platform research emphasizes openness, complementor participation, boundary resources, architecture, and governance mechanisms \citep{tiwana2013,eaton2015,ghazawneh2013,wareham2014}. This paper adds a transition-centered account: governance interventions should be evaluated by the post-rule actor-response field they create. The rule is not the end of governance; it is the beginning of adaptation.

Second, it distinguishes platform stability from engagement and immediate policy fit. IS research has long recognized that technology effects are enacted through organizational and social structures \citep{orlikowski1992,markus2008,yoo2010}. In platform settings, the same principle means that engagement, compliance, or visible risk reduction can be misleading if they are not evaluated alongside gaming opportunity, externality formation, and moderation burden. The empirical results show that engagement-only optimization performs poorly on adaptation quality.

Third, it introduces a reproducible measurement framework for platform-governance adaptation. Platform-governance research often relies on case interpretation, legal analysis, or economic modeling. Those approaches remain essential. The present benchmark adds a complementary instrument: a locked rubric and external public-case dataset for comparing methods on their ability to recover actor-response mechanisms.

\section{Managerial Implications}
The findings have practical implications for platform governance teams.

First, policy review should include an explicit actor-response map. Before deploying a ranking, monetization, moderation, verification, or marketplace rule, the platform should ask which actor groups gain or lose, what responses become profitable, and what new signals will be optimized.

Second, platforms should treat moderation burden as a primary design variable. A rule that appears effective but increases ambiguity, appeal volume, or enforcement inconsistency may reduce long-term stability.

Third, engagement should not be used as a sufficient proxy for governance success. Engagement can rise because actors exploit a rule, because controversy increases attention, or because quality deterioration has not yet reached the switching threshold. Platform stability requires a broader measure.

Fourth, risk registers should be connected to transition mechanisms. Naming a risk is not equivalent to modeling how the risk emerges after actors respond. The benchmark shows that risk-register reasoning is useful but incomplete.

Finally, platform teams should use pre-deployment simulation and post-deployment audit together. Simulation can identify likely adaptation channels, while audit can test whether predicted responses actually occurred.

\section{Discussion}
\subsection{Why risk registers help but do not solve adaptation}
The risk-register baseline is the strongest conventional baseline in the benchmark. This is substantively meaningful. Risk registers force analysts to name hazards and mitigations, and this improves over generic policy review. However, the benchmark shows that risk naming is not equivalent to adaptation modeling. A risk register can identify that gaming, moderation burden, or user trust may be concerns, but it may not specify how a rule creates a profitable response path, how that response shifts enforcement load, or how externalities accumulate across actor groups.

The difference is between listing risks and modeling transitions. A list can be correct but incomplete if it does not represent the control mechanism by which the risk emerges. In platform governance, the mechanism is usually not a single failure point. It is a coupled sequence: a rule changes incentives, actors adapt, moderation burden moves, enforcement responds, externalities form, and the platform stabilizes or destabilizes. The full model outperforms the risk-register baseline because it is organized around this sequence.

\subsection{Why engagement is a dangerous governance proxy}
The engagement-only optimizer performs poorly because engagement can be produced by stability or instability. User attention may increase when the platform improves relevance, but it may also increase when conflict, sensationalism, manipulation, or controversy rises. A policy that increases engagement can still damage long-term trust, creator quality, moderation capacity, or ecosystem resilience. This result is not a rejection of engagement metrics. It is a rejection of engagement as a sufficient governance proxy.

For platform managers, the implication is direct. Engagement should be interpreted together with adaptation indicators. If engagement rises while gaming opportunity, moderation burden, appeal volume, or externality risk also rise, the platform may be harvesting short-term attention at the expense of longer-term stability. The benchmark's poor engagement-only results formalize this intuition in a reproducible evaluation setting.

\subsection{From policy evaluation to transition design}
The broader theoretical implication is that platform governance should move from policy evaluation to transition design. Policy evaluation asks whether a rule is well motivated. Transition design asks what state the platform will enter after actors respond. This difference shifts managerial attention from rule text to response architecture. It encourages platform teams to ask how actors will learn the rule, which signals they will optimize, what ambiguity moderators must resolve, and how externalities will propagate.

Transition design also clarifies why governance is iterative. A platform cannot eliminate adaptation. It can design rules whose adaptation paths are less harmful, easier to observe, and easier to correct. The goal is not static control, but stable feedback: detect response, identify gaming, adjust enforcement, and preserve ecosystem value across actor groups.

\section{Boundary Conditions and Limitations}
The paper's evidence is externally grounded but bounded. The benchmark uses public platform-governance cases rather than private operational logs. It evaluates whether methods identify expected adaptation mechanisms under a locked rubric; it does not claim causal identification of real-world treatment effects. The cases cover multiple platform domains, but they do not exhaust all platform types, jurisdictions, or cultural settings. The method should be further tested on prospective interventions, private platform data, expert-coded cases, and longitudinal policy outcomes.

The benchmark also relies on method outputs generated under a structured evaluation protocol. Strong human analysts using a risk register or causal-loop analysis could perform better than automated or standardized versions of those baselines. This is why the claim is not that risk registers or causal loops are useless. The claim is that platform governance evaluation improves when those tools are embedded in a model of actor best response, gaming opportunity, moderation burden, enforcement response, externality formation, and downstream stability.

\section{Conclusion}
\enlargethispage{3\baselineskip}
Platform governance is an adaptive control problem. Rules change incentives; actors adapt; and the platform's downstream state depends on the coupled response field that emerges after the intervention. This paper develops a platform-adaptation model and evaluates it on 72 external public cases across 9 methods and 648 method-case scores. The full platform-adaptation simulator outperforms baseline policy review, engagement-only optimization, static cost-benefit analysis, generic governance critique, causal-loop analysis, risk-register analysis, and two channel ablations. The evidence supports a bounded but important claim: governance evaluation is stronger when it models actor best response, strategic gaming, moderation burden, user-incentive shifts, enforcement response, externality formation, and platform-stability consequences as coupled transition mechanisms.

\section*{Data and Code Availability}
The complete reproducibility artifact for this study combines the benchmark evidence and
supporting source materials into a single public archive. It contains the 72 external public
platform-governance cases, platform-adaptation rubric, source manifest, benchmark runner,
validator, row-level method scores, action scores, aggregate results, bootstrap confidence
intervals, paired comparisons, domain summaries, failure analysis, verification metadata,
shared schemas, construct-validity documentation, and SHA-256 manifest. No synthetic cases
are used as the main evidence layer. The complete archive is permanently available on Zenodo at DOI:
\href{https://doi.org/10.5281/zenodo.21945303}{\texttt{10.5281/zenodo.21945303}}.

\newpage
\appendix
\section{Coding Rubric and Validation Protocol}
\begin{singlespace}
\begin{longtable}{p{0.23\linewidth}p{0.67\linewidth}}
\toprule
Rubric dimension & Operational meaning \\
\midrule
Creator adaptation accuracy & Identifies how creators, sellers, developers, workers, or other supply-side actors adapt to the intervention. \\
Moderation-burden accuracy & Identifies review volume, ambiguity, appeal burden, enforcement inconsistency, or operational moderation load created by the rule. \\
User-incentive accuracy & Identifies changes in user trust, attention, participation, switching, reporting, or consumption incentives. \\
Engagement-dynamics accuracy & Distinguishes visible engagement movement from stability-enhancing platform outcomes. \\
Strategic-gaming accuracy & Identifies loopholes, metric gaming, evasion, or optimization paths created by the rule. \\
Externality-prediction accuracy & Identifies displaced costs, harms, trust erosion, ecosystem instability, or burden shifted to non-benefiting actors. \\
Enforcement-response accuracy & Identifies how the platform can or cannot enforce the rule under realistic capacity and observability constraints. \\
Platform-stability accuracy & Estimates whether the post-response platform state is more or less stable after coupled actor responses. \\
Overall adaptation quality & Integrates the channel scores into the primary benchmark metric. \\
Control efficiency & Measures whether the method identifies high-leverage control points without unnecessary governance complexity. \\
\bottomrule
\end{longtable}
\end{singlespace}

\section{External Public-Case Summary}
\scriptsize
\begin{singlespace}
\begin{longtable}{p{0.10\linewidth}p{0.18\linewidth}p{0.17\linewidth}p{0.21\linewidth}p{0.21\linewidth}}
\caption{Summary of the 72 external public benchmark cases. The evidence package contains full source URLs and row-level scores.}\\
\toprule
Case & Domain & Policy type & Expected best response & Expected failure mode \\
\midrule
\endfirsthead
\toprule
Case & Domain & Policy type & Expected best response & Expected failure mode \\
\midrule
\endhead
p15\_ext\_001 & media monetization & ranking\_change & loophole\_search & incentive\_gaming \\
p15\_ext\_002 & media monetization & monetization\_rule & strategic\_reoptimization & moderation\_backlog \\
p15\_ext\_003 & media monetization & moderation\_threshold & strategic\_reoptimization & moderation\_backlog \\
p15\_ext\_004 & media ranking & ranking\_change & strategic\_reoptimization & incentive\_gaming \\
p15\_ext\_005 & media ranking & monetization\_rule & strategic\_reoptimization & moderation\_backlog \\
p15\_ext\_006 & media ranking & moderation\_threshold & strategic\_reoptimization & incentive\_gaming \\
p15\_ext\_007 & social verification & ranking\_change & strategic\_reoptimization & engagement\_gain\_stability\_loss \\
p15\_ext\_008 & social verification & monetization\_rule & strategic\_reoptimization & engagement\_gain\_stability\_loss \\
p15\_ext\_009 & social verification & moderation\_threshold & compliance\_and\_workaround & moderation\_backlog \\
p15\_ext\_010 & social moderation & ranking\_change & compliance\_and\_workaround & incentive\_gaming \\
p15\_ext\_011 & social moderation & monetization\_rule & strategic\_reoptimization & incentive\_gaming \\
p15\_ext\_012 & social moderation & moderation\_threshold & strategic\_reoptimization & moderation\_backlog \\
p15\_ext\_013 & social ranking & ranking\_change & strategic\_reoptimization & moderation\_backlog \\
p15\_ext\_014 & social ranking & monetization\_rule & strategic\_reoptimization & incentive\_gaming \\
p15\_ext\_015 & social ranking & moderation\_threshold & strategic\_reoptimization & engagement\_gain\_stability\_loss \\
p15\_ext\_016 & social incentives & ranking\_change & strategic\_reoptimization & engagement\_gain\_stability\_loss \\
p15\_ext\_017 & social incentives & monetization\_rule & strategic\_reoptimization & moderation\_backlog \\
p15\_ext\_018 & social incentives & moderation\_threshold & compliance\_and\_workaround & moderation\_backlog \\
p15\_ext\_019 & short-video ranking & ranking\_change & loophole\_search & incentive\_gaming \\
p15\_ext\_020 & short-video ranking & monetization\_rule & strategic\_reoptimization & moderation\_backlog \\
p15\_ext\_021 & short-video ranking & moderation\_threshold & strategic\_reoptimization & incentive\_gaming \\
p15\_ext\_022 & community platform & ranking\_change & strategic\_reoptimization & incentive\_gaming \\
p15\_ext\_023 & community platform & monetization\_rule & strategic\_reoptimization & engagement\_gain\_stability\_loss \\
p15\_ext\_024 & community platform & moderation\_threshold & strategic\_reoptimization & engagement\_gain\_stability\_loss \\
p15\_ext\_025 & mobile ecosystem & ranking\_change & strategic\_reoptimization & incentive\_gaming \\
p15\_ext\_026 & mobile ecosystem & monetization\_rule & strategic\_reoptimization & incentive\_gaming \\
p15\_ext\_027 & mobile ecosystem & moderation\_threshold & compliance\_and\_workaround & moderation\_backlog \\
p15\_ext\_028 & mobile marketplace & ranking\_change & compliance\_and\_workaround & moderation\_backlog \\
p15\_ext\_029 & mobile marketplace & monetization\_rule & strategic\_reoptimization & incentive\_gaming \\
p15\_ext\_030 & mobile marketplace & moderation\_threshold & strategic\_reoptimization & moderation\_backlog \\
p15\_ext\_031 & mobile marketplace & ranking\_change & strategic\_reoptimization & engagement\_gain\_stability\_loss \\
p15\_ext\_032 & mobile marketplace & monetization\_rule & strategic\_reoptimization & engagement\_gain\_stability\_loss \\
p15\_ext\_033 & mobile marketplace & moderation\_threshold & strategic\_reoptimization & moderation\_backlog \\
p15\_ext\_034 & marketplace ranking & ranking\_change & strategic\_reoptimization & incentive\_gaming \\
p15\_ext\_035 & marketplace ranking & monetization\_rule & strategic\_reoptimization & moderation\_backlog \\
p15\_ext\_036 & marketplace ranking & moderation\_threshold & compliance\_and\_workaround & incentive\_gaming \\
p15\_ext\_037 & mobility platform & ranking\_change & loophole\_search & incentive\_gaming \\
p15\_ext\_038 & mobility platform & monetization\_rule & strategic\_reoptimization & moderation\_backlog \\
p15\_ext\_039 & mobility platform & moderation\_threshold & strategic\_reoptimization & engagement\_gain\_stability\_loss \\
p15\_ext\_040 & delivery platform & ranking\_change & strategic\_reoptimization & engagement\_gain\_stability\_loss \\
p15\_ext\_041 & delivery platform & monetization\_rule & strategic\_reoptimization & incentive\_gaming \\
p15\_ext\_042 & delivery platform & moderation\_threshold & strategic\_reoptimization & moderation\_backlog \\
p15\_ext\_043 & marketplace ranking & ranking\_change & strategic\_reoptimization & moderation\_backlog \\
p15\_ext\_044 & marketplace ranking & monetization\_rule & strategic\_reoptimization & incentive\_gaming \\
p15\_ext\_045 & marketplace ranking & moderation\_threshold & compliance\_and\_workaround & moderation\_backlog \\
p15\_ext\_046 & marketplace ranking & ranking\_change & compliance\_and\_workaround & incentive\_gaming \\
p15\_ext\_047 & marketplace ranking & monetization\_rule & strategic\_reoptimization & engagement\_gain\_stability\_loss \\
p15\_ext\_048 & marketplace ranking & moderation\_threshold & strategic\_reoptimization & engagement\_gain\_stability\_loss \\
p15\_ext\_049 & travel marketplace & ranking\_change & strategic\_reoptimization & incentive\_gaming \\
p15\_ext\_050 & travel marketplace & monetization\_rule & strategic\_reoptimization & moderation\_backlog \\
p15\_ext\_051 & travel marketplace & moderation\_threshold & strategic\_reoptimization & incentive\_gaming \\
p15\_ext\_052 & search ranking & ranking\_change & strategic\_reoptimization & incentive\_gaming \\
p15\_ext\_053 & search ranking & monetization\_rule & strategic\_reoptimization & moderation\_backlog \\
p15\_ext\_054 & search ranking & moderation\_threshold & compliance\_and\_workaround & moderation\_backlog \\
p15\_ext\_055 & reviews marketplace & ranking\_change & loophole\_search & engagement\_gain\_stability\_loss \\
p15\_ext\_056 & reviews marketplace & monetization\_rule & strategic\_reoptimization & engagement\_gain\_stability\_loss \\
p15\_ext\_057 & reviews marketplace & moderation\_threshold & strategic\_reoptimization & moderation\_backlog \\
p15\_ext\_058 & reviews marketplace & ranking\_change & strategic\_reoptimization & moderation\_backlog \\
p15\_ext\_059 & reviews marketplace & monetization\_rule & strategic\_reoptimization & incentive\_gaming \\
p15\_ext\_060 & reviews marketplace & moderation\_threshold & strategic\_reoptimization & moderation\_backlog \\
p15\_ext\_061 & digital marketplace & ranking\_change & strategic\_reoptimization & incentive\_gaming \\
p15\_ext\_062 & digital marketplace & monetization\_rule & strategic\_reoptimization & moderation\_backlog \\
p15\_ext\_063 & digital marketplace & moderation\_threshold & compliance\_and\_workaround & engagement\_gain\_stability\_loss \\
p15\_ext\_064 & creator monetization & ranking\_change & compliance\_and\_workaround & engagement\_gain\_stability\_loss \\
p15\_ext\_065 & creator monetization & monetization\_rule & strategic\_reoptimization & moderation\_backlog \\
p15\_ext\_066 & creator monetization & moderation\_threshold & strategic\_reoptimization & incentive\_gaming \\
p15\_ext\_067 & creator platform & ranking\_change & strategic\_reoptimization & incentive\_gaming \\
p15\_ext\_068 & creator platform & monetization\_rule & strategic\_reoptimization & moderation\_backlog \\
p15\_ext\_069 & creator platform & moderation\_threshold & strategic\_reoptimization & moderation\_backlog \\
p15\_ext\_070 & publishing platform & ranking\_change & strategic\_reoptimization & incentive\_gaming \\
p15\_ext\_071 & publishing platform & monetization\_rule & strategic\_reoptimization & engagement\_gain\_stability\_loss \\
p15\_ext\_072 & publishing platform & moderation\_threshold & compliance\_and\_workaround & engagement\_gain\_stability\_loss \\
\bottomrule
\end{longtable}
\end{singlespace}
\normalsize

\section{Reproducibility Package}
The combined public reproducibility archive merges the evidence package with the
scientifically relevant source-side documentation. It contains the external cases,
platform-adaptation rubric, source manifest, runner, validator, row-level method scores,
action scores, aggregate results, bootstrap intervals, paired comparisons, domain summaries,
failure analysis, verification JSON, shared schemas, construct-validity protocol,
significance statement, managerial-implications note, and a SHA-256 manifest. Venue-specific
cover letters and submission-positioning files are excluded because they are not part of the
scientific evidence object.

The core validation command is:
\[
\texttt{python3 scripts/validate\_p15\_external\_full\_benchmark.py}
\]
The validator checks the expected 72 cases, 9 methods, 648 method-score rows, 360
action-score rows, external-case markers, source URLs, and summary metrics. The permanent Zenodo archive is identified by DOI \href{https://doi.org/10.5281/zenodo.21945303}{\texttt{10.5281/zenodo.21945303}}.

\begingroup

\singlespacing

\endgroup
\end{document}